\documentclass[10pt, conference, compsocconf]{IEEEtran}
\usepackage[pdftex]{graphicx}
\usepackage[pagebackref=true,breaklinks=true,colorlinks,bookmarks=false]{hyperref}
\usepackage{times}
\usepackage{epsfig}
\usepackage{graphicx}
\usepackage{amsmath}
\usepackage{amssymb}
\usepackage{subfig}
\usepackage{bbm}
\usepackage[belowskip=0pt,aboveskip=0pt]{caption}
\newcommand{\fig}[1]{Fig.~\ref{#1}} 
\usepackage{tabularx}
\usepackage{array}
\usepackage{multirow}
\usepackage{booktabs}
\usepackage{color, colortbl}
\usepackage{authblk}
\usepackage{float}
\usepackage{hhline}
\usepackage{bm}
\definecolor{Gray}{gray}{0.9}
\begin{document}
\title{A Hierarchical Deep Architecture and Mini-Batch Selection Method For Joint Traffic Sign and Light Detection}

\author[]{Alex D. Pon}
\author[]{Oles Andrienko}
\author[]{Ali Harakeh}
\author[]{Steven L. Waslander}
\affil[] {
Mechanical and Mechatronics Engineering Department
\\
University Of Waterloo
\\
Waterloo, ON, Canada
}
\affil[]{\textit{apon@uwaterloo.ca, oandrien@uwaterloo.ca, www.aharakeh.com, stevenw@uwaterloo.ca}}

\maketitle

\begin{abstract}

Traffic light and sign detectors on autonomous cars are integral for road scene perception. The literature is abundant with deep learning networks that detect either lights or signs, not both, which makes them unsuitable for real-life deployment due to the limited graphics processing unit (GPU) memory and power available on embedded systems. The root cause of this issue is that no public dataset contains both traffic light and sign labels, which leads to difficulties in developing a joint detection framework. We present a deep hierarchical architecture in conjunction with a mini-batch proposal selection mechanism that allows a network to detect both traffic lights and signs from training on separate traffic light and sign datasets. Our method solves the overlapping issue where instances from one dataset are not labelled in the other dataset. We are the first to present a network that performs joint detection on traffic lights and signs. We measure our network on the Tsinghua-Tencent 100K benchmark for traffic sign detection and the Bosch Small Traffic Lights benchmark for traffic light detection and show it outperforms the existing Bosch Small Traffic light state-of-the-art method. We focus on autonomous car deployment and show our network is more suitable than others because of its low memory footprint and real-time image processing time. Qualitative results can be viewed at \href{https://youtu.be/_YmogPzBXOw}{https://youtu.be/\_YmogPzBXOw}.
\end{abstract}


\IEEEpeerreviewmaketitle

%------------------------------------------------------
\section{Introduction} \label{introduction}

Recent progress in deep learning has led to a proliferation of deep learning networks on object detection dataset leaderboards. These leaderboards, however, often do not account for the graphics processing unit (GPU) memory and power required, nor the run-time speed of these networks. These considerations are important in autonomous driving; cost and power limitations make it unsustainable to have multiple onboard GPUs to support deep neural networks, and real-time detection speeds are essential for safety. Ideally, a single network would perform multiple tasks to preserve GPU memory and power.

Unfortunately, in the domain of traffic light and sign detection, networks often only detect traffic lights \textit{or} signs because no public dataset includes both traffic light \textit{and} sign labels. The COCO dataset \cite{COCO} comes close with labelled traffic lights and signs, but it does not differentiate traffic light states and only labels \textit{stop} signs. As such, to have state-of-the-art traffic light and sign detection on an autonomous car, currently two separate networks are needed. This work takes a step towards joint traffic light and sign detection with the motivation of minimizing GPU memory and power required for both tasks.

Since no public dataset contains the labels for both traffic lights and signs, it is logical to try to train on a combination of datasets. One difficulty in combining datasets is that the network is required to approximate a more complex function, so lower performance is expected than if individual networks were used. To minimize this loss of performance we propose an implicit, hierarchical neural network architecture that exploits the characteristic that traffic lights only differ by their state and that traffic signs share similar features. The network determines the global class of a proposal, which we define as either traffic light, sign or background. The network also determines the subclass of each proposal, but only proposals that correctly identify the global class are considered in the loss function. 

Another difficulty in combining datasets is overlapping; both datasets contain unlabeled instances of the object of interest from the other dataset. Overlapping will confuse the network during training when it correctly classifies the unlabeled instances. For instance, a network could identify a traffic light in the traffic sign dataset, but since it is unlabelled it would be penalized for the correct detection. To overcome the described issue we propose a mechanism referred to as the \textit{background threshold}. The background threshold alters the standard region proposal network (RPN) \cite{ren2015} mini-batch process by requiring proposals labeled as the background class to have a minuscule overlap with a ground truth box. This method reduces the likelihood of an unlabeled object of interest considered as the background because it exploits the characteristic that traffic signs and lights are naturally separated in physical space.

The contributions of this paper are as follows:
\begin{itemize}
    \item We present an implicit, hierarchical approach to traffic light and sign detection that classifies objects according to their global categories first and subclasses second.
    \item We present a novel approach to train on two overlapping datasets. Our approach does not add any new parameters and reduces the probability of confusing the background class with unlabeled objects of interest from the datasets.
    \item We are the first to present a network that performs joint traffic light and sign detection. Our architecture is suitable for autonomous car deployment because it saves GPU memory by solving two tasks and has real-time detection speeds. We test our single network on the Bosch Small Traffic Light dataset \cite{behrendt2017} and the Tsinghua-Tencent 100K Traffic Sign dataset \cite{tencent} and show it outperforms the existing Bosch dataset state-of-the-art.
\end{itemize}

The rest of the paper is structured as follows. Section \ref{related_work} provides an overview of the state-of-the-art in traffic light and sign detection. Section \ref{prob_form} provides the problem formulation and our proposed solutions. Section \ref{experiments} discusses the results on the two test datasets. Lastly, we conclude the paper with Section \ref{conclusion}.

%------------------------------------------------------
\section{Related Work} \label{related_work}

\subsection{Traffic Sign and Traffic Light Datasets}
The existing public traffic light datasets include the LaRA dataset \cite{lara}, LISA dataset \cite{lisa1, lisa2}, and the Bosch dataset. The LaRA dataset has low resolution images and a \textit{temporal matching} evaluation procedure that makes it inadequate for training our deep architecture. There are also several flaws in the LISA dataset. Lee et al. \cite{lee} noted annotations for small traffic lights are missing. Furthermore, there is inconsistency in which images and classes are used for evaluation, which makes it difficult to have comparisons; \cite{lee,li2017a} only evaluate using green and red lights and \cite{lee} uses part of the designated training set as their test set. The recently proposed Bosch dataset was chosen for training and evaluation because the small traffic lights it includes makes it challenging and the evaluation procedure is clear because of their available test set.

There are many European traffic sign datasets. The most common is the German Traffic Sign Detection Benchmark (GTSDB) \cite{GTSDB} and the German Traffic Sign Recognition Benchmark (GTSRB) \cite{GTSRB}, which test object detection and classification, respectively. GTSRB is not representative of a real driving scenario as it contains crops of traffic signs, whereas real traffic signs are only a small component of the image in road scenes. Moreover, the GTSDB only contains 4 classes allowing multiple unpublished methods to achieve $100\%$ accuracy and precision.  We have chosen to use the Tsinghua-Tencent 100K dataset because it is significantly more challenging; there are 45 classes of signs, and the images are not cropped to include only the traffic sign extent. Moreover, the recent attempts \cite{tencent, perceptual_gan, meng} on this dataset and standard evaluation procedure allows our work to be compared to the state-of-the-art.

\begin{figure*}[t]
\begin{center}
\includegraphics[width=\textwidth]{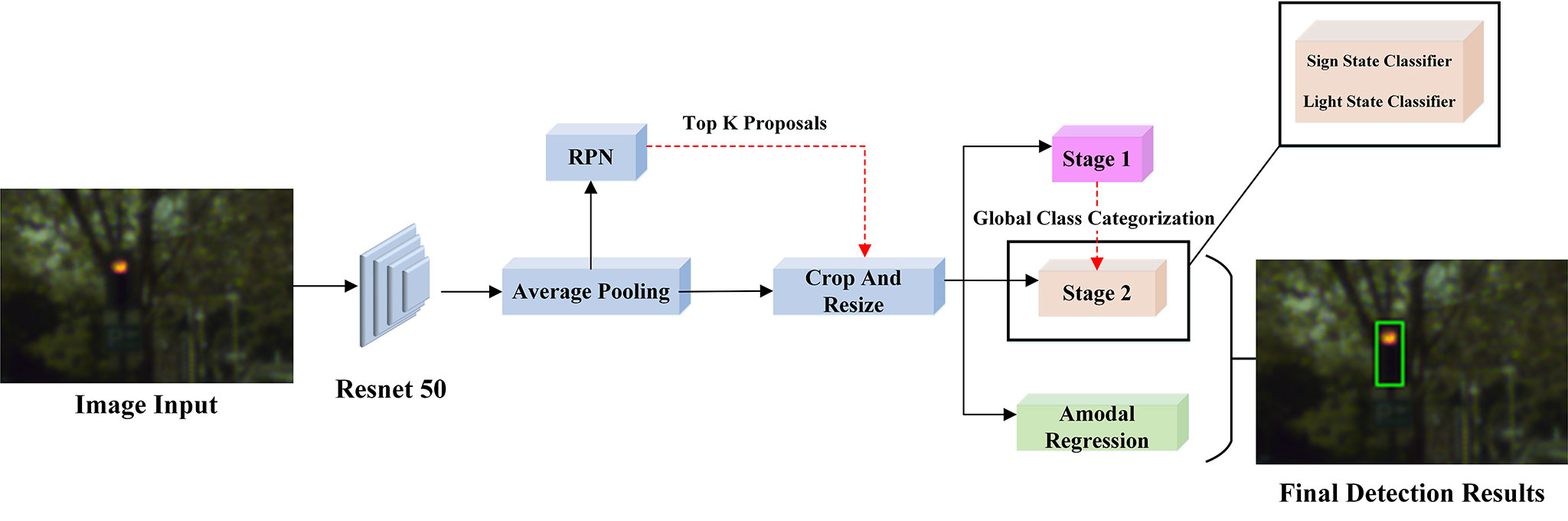}
\end{center}
\caption{The hierarchical architecture used to jointly detect traffic signs and lights.}
\label{arch}
\end{figure*}

\subsection{Traffic Light Detectors}
There are two main categories for traffic light detection algorithms: image processing based models and learning based models. Driven by intuition, initial traffic light detection approaches often used image processing based models, which rely on traffic light shape and color as cues for detection. Franke et al. \cite{franke1999} and Lindner et al. \cite{lindner} classify the color of each pixel and then use connected component analysis for segmentation to determine regions of interest. Other image processing approaches \cite{masakoomachi2009,yehushen2009} often normalize the color space of the image and then identify and group pixels that exceed specified thresholds. These groups are then further processed by imposing shape constraints to find the traffic lights. In addition, in \cite{lindner,fairfield2011}, prior maps were used to minimize false positives by providing traffic light location and state information. 

Image processing based methods have several advantages. These methods do not require training data or suffer from overfitting, and they tend to work well when tailored for specific scenarios. Due to the rigidness of these deterministic methods, however, they are prone to break under slight variability. As an example, algorithms tailored to handle round traffic lights struggle when the traffic light states include arrows. Moreover, methods that rely on prior maps are limited in areas without mapping information.

Learning based models can overcome many of the limitations incurred by image processing based models as they can be trained on a broad set of traffic lights that include arrows and lights in various environments. Learning based models have also been proven to outperform image processing based models in traffic light detection. In \cite{jensen2015, lisa1}, ACF detectors outperformed several image processing based detectors on the LISA dataset. Despite the advantages of learning based traffic light detectors, their usage is relatively new so only a few other methods have been implemented. In \cite{behrendt2017}, YOLO \cite{redmon2015} is used to detect traffic lights and a separate convolutional neural network (CNN) classifies the traffic light states. In \cite{jensen}, YOLO2 \cite{redmon2016} is used on the LISA dataset. We introduce a modified Faster R-CNN \cite{ren2015} architecture that outperforms the current state-of-the-art on the Bosch dataset.

\subsection{Traffic Sign Detectors}
Traffic sign and traffic light detection research have followed a similar trend. Classical traffic sign detection methods rely on color and shape information \cite{escalera, benallal, ruta}. Learning based methods became prevalent after the emergence of AlexNet \cite{Krizhevsky} and other deep learning networks. Using learning based methods is logical for traffic sign detection because signs have variability in color and shape, making it difficult to create hand-crafted features that generalize well for all signs. In addition, similarly in traffic light detection, learning based methods provide an opportunity to solve illumination changes and varying viewpoints. Examples of deep learning traffic sign detection approaches include multi-scale CNN \cite{sermanet}, Support Vector Machines (SVMs) to classify into global classes coupled with a CNN to further classify to finer categorization \cite{yang}, a CNN with diluted convolutions \cite{aghdam}, and a CNN with a Generative Adversarial Network (GAN) that enhances small images \cite{perceptual_gan}. Our approach is the first to merge traffic sign and traffic light detection without a large loss in performance in either one of the global class categorization.

\subsection{Hierarchical Object Detection}
Using a hierarchical scheme to classify an object into a generic category then a specific category has been studied in works including \cite{meger, hoai}. We extend these works by applying a hierarchy to joint traffic light and sign classification.

\begin{figure}[b]
\begin{center}
\includegraphics[width=\columnwidth]{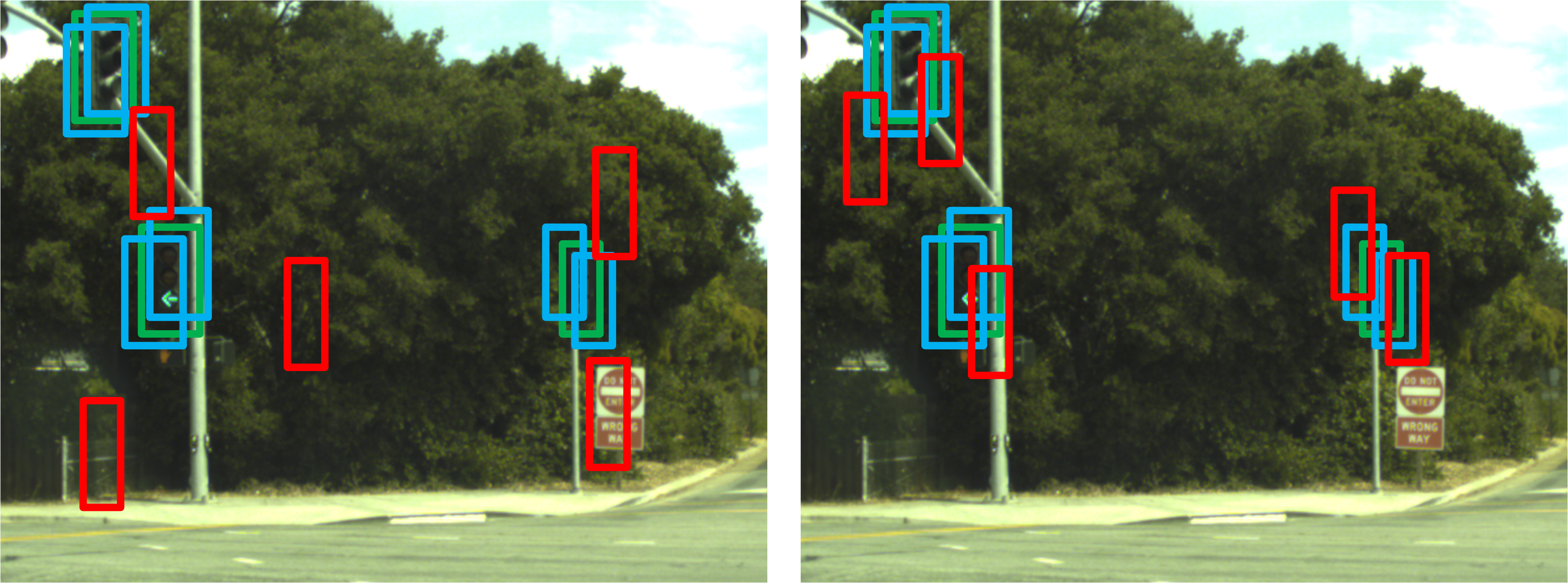}
\end{center}
\caption{An image from the Bosch dataset \cite{behrendt2017}. Ground truth is shown in \textit{green}, while the positive and negative members of the mini-batch are shown in \textit{blue} and \textit{red} respectively. \textbf{Left}: The standard mini-batch selection process causes some negative proposals to contain the unlabeled object of interest. \textbf{Right}: Our mini-batch selection method minimizes negative selection of non-labeled objects by sampling near the labeled ground truth box. }
\label{mini_batch}
\end{figure}

\section{Problem Formulation and Proposed Solutions} \label{prob_form}
This section describes the problem formulation of 2D object detection, the difficulties induced by combining two datasets, and our proposed approaches to solve both of these problems.

\subsection{2D Object Detection}
\noindent\textbf{Problem Formulation:} 
Let $f$ be a function that maps a $M \times N$ image $I$ to a set $\bm{O}$, which represents all objects of interest in a dataset. Each object $o_i$ is indexed by $i \in \bm{O}$ and is parameterized by $\{x_1,y_1,x_2,y_2\} \in \mathbb{R}^4$, the coordinates of the top left and bottom right corners of the \textit{tightest} bounding box around the object, and $c \in \bm{K}=\{1, \ldots, K\} \subseteq \mathbb{Z}$ a discrete integer variable that defines the class of an object from a choice of $K$ classes.
The problem is thus defined as:   
\begin{equation}
 \begin{aligned}
 \mbox{Find} \ f:& \ \mathbb{R}^{M\times N} \rightarrow (\mathbb{R}^4 \times K)^I  \\
 f(\bm{I}&)=\bm{O} \\ 
 \bm{O}: \ &\{o_i=\{ x_{1,i},x_{2,i},y_{1,i},y_{2,i},c_i\}| \ \forall i \in \bm{O}: \\
 & 0 \leq x_1, x_2 \leq M\\
 & 0\leq y_1, y_2 \leq N\\
 & x_1 < x_2 \\
 & y_1 < y_2 \\
 & c \in \bm{K}\}.
 \end{aligned}
 \label{probform}
\end{equation}
Solving the generalized 2D object detection problem is out of the scope of this paper, so we restrict ourselves to $K=50$ classes that describe traffic lights and signs.
\\

\noindent\textbf{Proposed Solution:} 
We tackle the problem of 2D object detection with a hierarchical deep architecture that jointly regresses the bounding box coordinates $\{x_1,y_1,x_2,y_2\}$ and determines the class $c$. Our hierarchical approach, shown in \fig{arch}, is built upon the ResNet-50 \cite{He} version of Faster R-CNN. We use Faster R-CNN's RPN to generate $j$ amodal region proposals. Each region proposal is parameterized by a feature vector $\bm{h}$ extracted from the average pool feature map via a crop and re-size operation. We model the function in Eq. \ref{probform} as the following:

\begin{equation}
\begin{aligned}
f(\bm{I}) &= \begin{bmatrix}
g(\bm{h})\\ k(\bm{h}) 
\end{bmatrix},
\end{aligned}
\label{det_sol}
\end{equation}
where $g:\mathbb{R}^{2048} \rightarrow \mathbb{R}^4$ and $k:\mathbb{R}^{2048} \rightarrow \mathbb{Z}$. The function $g$ is modeled by the regression layer of Faster R-CNN and estimates the extent of a bounding box regardless of its class. The function $k$, on the other hand, is different from the classification layer used by Faster R-CNN and is modeled as a hierarchical classifier with two stages. The first stage classifies the global category of the region proposal as either traffic light, sign, or background. The second stage then classifies the subclass of the region proposal \textit{only} if the first stage predicts a traffic light or sign. Traffic light subclasses include red, green, yellow, etc., while traffic sign subclasses include stop, yield, no parking, etc. We use a hierarchical architecture to exploit that similar features are shared between the subclasses of traffic lights and between the subclasses of traffic signs; traffic lights only differ by their state and traffic signs are generally constrained to specific shapes and colors. \\

\noindent\textbf{Implementation Details:}  
Instead of explicitly forcing hierarchy in the neural network layers, we realize the proposed hierarchical classifier by adding a global category classification layer parallel to the second stage classifier as shown in \fig{arch} and modify the classification network loss function. The original classification loss function of Faster R-CNN is as follows:

\begin{equation}
\begin{aligned}
L({p_i}, {t_i}) = &\frac{1}{N_{tot}}\sum_i L_{cls}(p_i, u_i) + \\ &\lambda \frac{1}{N_{+}} \sum_i p^*_i L_{loc}(t_i, v_i),
\end{aligned}
\label{loss_original}
\end{equation}
where $p_i$ and $t_i$ are the classification and regression outputs, and $u_i$ and $v_i$ are the classification and regression ground truth, respectively. The variable $p^*_i$ is an indicator of whether the region proposal has an intersection-over-union (IoU) greater than $0.5$ with any ground truth bounding box. Proposals that satisfy this requirement are referred to as \textit{positive}. The cross-entropy loss $L_{cls}$ is normalized by the total number of proposals $N_{tot}$, while $L_{loc}$ is the smooth L1 loss normalized by the number of positive proposals $N_{+}$. A factor $\lambda$ is used to balance the two losses. We modify the original loss function to account for our hierarchical architecture as follows:

\begin{equation}
\begin{aligned}
L({p_i},{t_i}) = & \frac{1}{N_{tot}}\sum_{i}L_{cls}(p_{i}, u_{i}) +\\&\frac{1}{N_{\dagger}}\sum_{i}p^{\dagger}_{i}L_{cls}(\hat{p_{i}}, \hat{u_{i}})+\\ &\lambda\frac{1}{N_{+}}\sum_{i} p^*_{i}L_{loc}(t_i,v_i).
\end{aligned}
\end{equation}
where $\hat{p_i}$ and $\hat{u_i}$ are subclass predictions and ground truth, respectively. In the first term, the classification cross-entropy loss is computed for all proposals based on the predicted global classes. Proposals are given a $p^{\dagger}_{i}$ value of 1 if their global class was classified correctly, or 0 if it was incorrect; therefore, proposals with an incorrect global class do not contribute to the subclass classification loss. The subclass classification loss is normalized by the total number of correctly identified global class proposals $N_{\dagger}$. The regression loss function remains the same. To compensate for the changed classification-to-regression loss ratio, $\lambda$ is doubled.

\subsection{Training On Two Overlapping Sets With Missing Labels}
\noindent\textbf{Problem Formulation:} 
Like before, let $\bm{O}$ be the set of labels for all traffic lights and signs. Let $\mathcal{L}$ be the set of frames with labels for traffic lights only --- a traffic light data set. Also, let $\mathcal{S}$ be the set of frames with labels for traffic signs only --- a traffic sign dataset. If we take $\mathcal{L} \cup \mathcal{S}$, we result in a dataset with unlabeled traffic lights in $\mathcal{S}$ and unlabeled traffic signs in $\mathcal{L}$; in other words, $\mathcal{L} \cup \mathcal{S} \subsetneq \bm{O}$. \fig{mini_batch} shows how these missing labels creates issues for the mini-batch selection process for any two-stage object detector. The problem is thus the following: train a function approximator that estimates $f(I)$ such that non-labeled objects of interest in the joint dataset do not degrade the estimation quality. \\

\begin{figure*}[t!]
    \begin{center}
    \includegraphics[width=\textwidth]{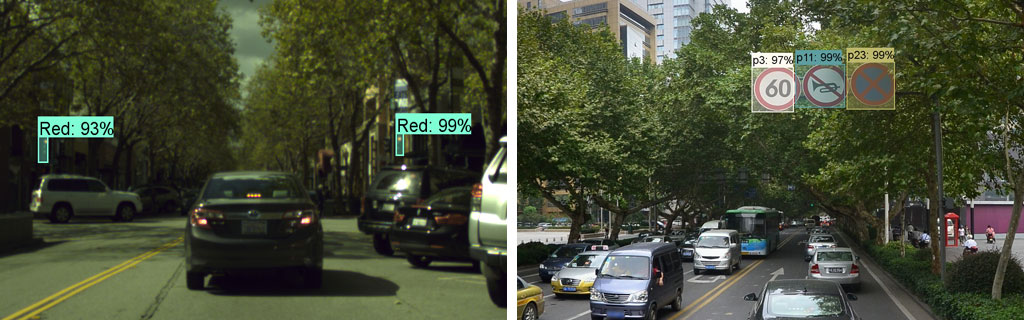}
    \end{center}
    \caption{Results from the Hierarchical + Threshold model on the Bosch dataset (left) and on the Tsinghua-Tencent dataset (right).}
    \label{sample_detection_bosch_tencent}
\end{figure*}

\begin{table*}[t]
\centering
\resizebox{\textwidth}{!}{
\begin{tabular}{cccc}
\midrule
\textbf{Method} & \textbf{Bosch mAP} & \textbf{Tsinghua-Tencent mAP} & \textbf{Total mAP}\\
\midrule
\midrule
Behrendt \cite{behrendt2017} Trained on Bosch &~0.40 & - & - \\
\midrule
Baseline Trained on Bosch &\textbf{0.53}& - & -\\
Baseline Trained on Tsinghua-Tencent 100K &-& \textbf{0.40}& -\\
Baseline Trained on Bosch and Tsinghua-Tencent 100K & 0.43 & 0.26 & 0.34 \\ 
\midrule
Hierarchical Model &0.45& 0.30 & 0.37\\
Background Threshold Model &0.41& 0.32 & 0.37\\
Hierarchical + Background Threshold Model & 0.46 & 0.31 & \textbf{0.38}\\ 
\bottomrule\\
\end{tabular}}
\caption{Comparison of the performance of our baselines and novel architectures on the Bosch and Tsinghua-Tencent datasets.}
\label{performance_table}
\end{table*}

\noindent\textbf{Proposed Solution:} Our proposed solution leverages Faster R-CNN's mini-batch selection mechanism. Let $\mathcal{B}$ be the set of all proposals and $\mathcal{G}$ be the set of ground truth bounding boxes in a frame $I$. In the original Faster R-CNN mini-batch selection, proposals are separated into a positive set $\mathcal{B}_{+}$ and a negative set $\mathcal{B}_{-}$ for loss computation according to:
\begin{equation}
\begin{aligned}
\mathcal{B}_{+} &= \{b_i| 0.7 \leq IoU(b_i,g_i) \leq 1\}, \\
\mathcal{B}_{-} &= \{b_i| IoU(b_i,g_i) \leq 0.3 \}, \\
\forall b_i& \in \mathcal{B}, g_i \in \mathcal{G}, 
\end{aligned}
\end{equation}
where $IoU$ is the intersection over union between $b_i$, an element of $\mathcal{B}$, and $g_i$, an element of $\mathcal{G}$. 

To solve the overlapping issue we enforce a lower IoU bound of $0.01$ for negative proposals, that we refer to as the \textit{background threshold}. By requiring this minuscule overlap, we reduce the likelihood of an unlabelled object of interest considered as the background because it exploits the characteristic that traffic signs and lights are naturally separated in physical space. A negative proposal is less likely to contain an unlabeled object if we sample close to the ground truth box of the labeled object. Although the validity of this prior is dependent on the dataset, we verified that only $\sim$7\% of images in the Tsinghua-Tencent dataset violate this assumption, and we expect this result to be similar for other traffic sign and light datasets. Our mini-batches are selected according to:
\begin{equation}
\begin{aligned}
\mathcal{B}_{+} &= \{b_i| 0.7 \leq IoU(b_i,g_i) \leq 1\}, \\
\mathcal{B}_{-} &= \{b_i| 0.01 \leq IoU(b_i,g_i) \leq 0.3 \}, \\
\forall b_i& \in \mathcal{B}, g_i \in \mathcal{G},
\end{aligned}
\end{equation}
which introduces no additional computation overhead and allows training on two overlapping training sets with missing target class labels.

\section{Experiments} \label{experiments}

We evaluate our proposed solutions described in Section \ref{prob_form} by performing experiments on the Bosch and Tsinghua-Tencent datasets. For training, we reduce the Tsinghua-Tencent dataset to 45 classes as done in \cite{tencent,perceptual_gan, meng}. We similarly limit the Bosch training dataset to 5 classes to avoid training on sparse classes. All of our novel models were trained on both the Bosch and Tsinghua-Tencent dataset and their results are shown in Table \ref{performance_table}, and sample detections are shown in \fig{sample_detection_bosch_tencent}.

\begin{figure*}[t]
    \begin{center}
    \includegraphics[width=\textwidth]{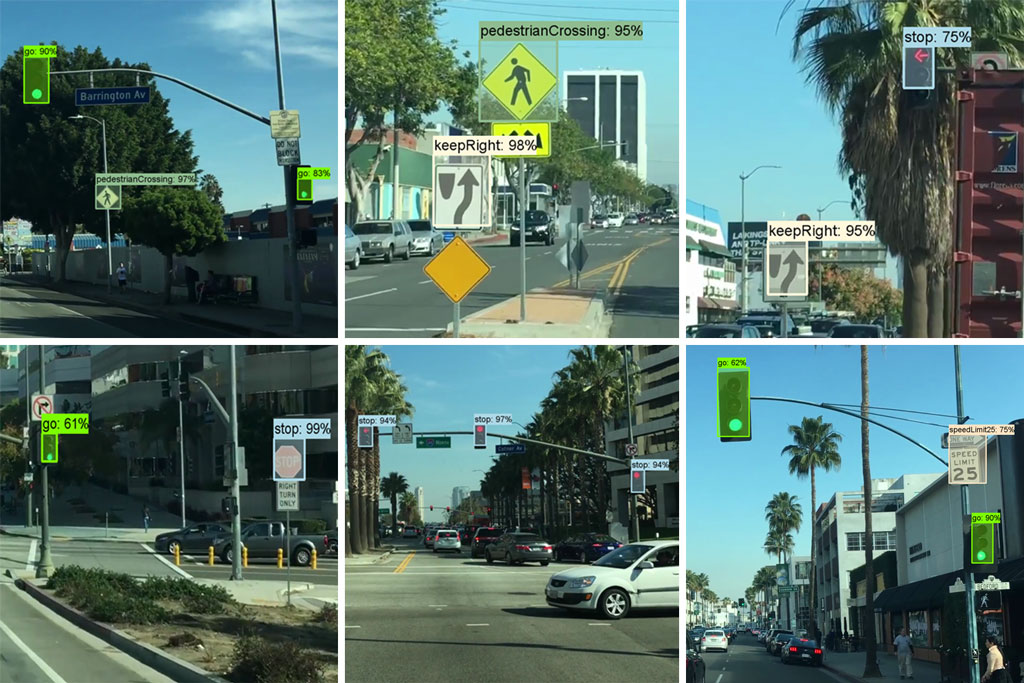}
    \end{center}
    \caption{Results of the Hierarchical + Threshold model on images from Los Angeles, United States \cite{la_video}. This model was trained on images from San Diego, United States (LISA Sign \cite{lisa_signs} and LISA Light datasets \cite{lisa1,lisa2}). This shows the model was able to generalize well to images from another data source and city.}
    \label{sample_detection_la}
\end{figure*}

\subsection{Implementation and Evaluation Procedure}

\noindent\textbf{Implementation}:  Baseline networks used the Tensorflow \cite{abadi2016} implementation of Faster R-CNN with ResNet-50, and all models were trained with stochastic gradient descent with momentum of 0.9 on a NVIDIA GeForce GTX 1080 Ti GPU. Following \cite{huang2016}, the maximum number of proposals generated by the RPN was set to 50. Other hyperparameters were unmodified from their default value. On average all models process one image in 0.015 seconds. \\

\noindent\textbf{Bosch Evaluation Procedure}: Following the procedure of \cite{behrendt2017}, precision-recall curves were used. We required our detections to have an IoU $\geq$ 0.5 with the ground truth boxes. We also compute mean average precision (mAP) which will allow for better future comparisons; \cite{behrendt2017} only provides precision-recall curves which makes it difficult to make quantitative comparisons. \\

\noindent\textbf{Tsinghua-Tencent Evaluation Procedure}: Following \cite{tencent}, we evaluate our approaches on accuracy and recall metrics on small (area $<$ $32^2$ pixels), medium ($32^2$ $<$ area $<$ $96^2$) and large (area $>$ $96^2$) objects used in the Microsoft COCO benchmark. Although better suited metrics exist for object detection such as mAP, average recall is sufficient for comparison due to the respective correlation with detection performance \cite{effective_detections}. With a minimum IoU threshold of $\geq$ 0.5, we evaluate over all classes to determine the final model performances with respect to the Tsinghua-Tencent test set. \\

\subsection{Comparison to Bosch Dataset State-of-the-Art}

Table \ref{performance_table} shows that the baseline network trained on the Bosch dataset outperformed the model in \cite{behrendt2017}. Behrendt et al. \cite{behrendt2017} provide a precision-recall curve created from their network that detects a generic traffic light class, and they provide a classification accuracy from their network that classifies the generic traffic light detections. We generously estimate their mAP as $0.40$ by measuring the area under their precision-recall curve and assuming a classification accuracy of $100\%$.  Our baseline model's mAP is 0.54. We mostly attribute the difference with YOLO's localization weakness; YOLO ``uses relatively coarse features for predicting bounding boxes'' \cite{redmon2015}, which could affect its ability in detecting small objects. In addition, all of our novel models outperformed \cite{behrendt2017}.

\subsection{Comparison to Tsinghua-Tencent Dataset State-of-the-Art}

In Table \ref{tencent_compare}, we provide a comparison with our best model trained on the Tsinghua-Tencent dataset and the state-of-the-art. Our model's low accuracy and recall can be attributed to the architecture's design for real-time performance; the motivation for our research is to implement a detection network on an autonomous vehicle, so we use a lightweight feature extractor and low region proposal count. The methods we compare to do not make the same prioritization and thus do not achieve real-time performance. Li et al. \cite{perceptual_gan} has a slow detection time of $0.6$ seconds per frame excluding proposal time because of their generative model, and \cite{meng} and \cite{tencent} do not provide inference time, but both state speed as a needed improvement for their models. Meng et al. \cite{meng} uses an expensive image pyramid and sliding window approach and \cite{tencent} uses the computationally intensive \textit{OverFeat} \cite{overfeat} framework.  Our model is able to perform inference at $0.015$ seconds per image, a 40X speedup over \cite{perceptual_gan}.

\begin{table}[t]
\centering
\resizebox{\columnwidth}{!}{
\begin{tabular}{ccccc}
\midrule
\textbf{} & \textbf{Small} & \textbf{Medium} & \textbf{Large} & \multicolumn{1}{l}{\textbf{Overall}} \\
\midrule
\midrule
Zhu \cite{tencent} (A)          & 82 & 91 & 91 & 88  \\
Zhu \cite{tencent} (R)          & 87 & 94 & 88 & 90  \\ 
\midrule
Li \cite{perceptual_gan} (A)    & 84 & 91 & 91 & 89 \\
Li \cite{perceptual_gan} (R)    & 89 & 96 & 89 & 91 \\ 
\midrule
Meng \cite{meng} (A)            & - & - & - & 90  \\
Meng \cite{meng} (R)            & - & - & - & 93  \\ 
\midrule
Ours (A)                        & 65 & 67 & 75 & 68 \\
Ours (R)                        & 24 & 54 & 70 & 44 \\
\bottomrule \\
\end{tabular}}
\caption{Accuracy (A) and recall (R) comparison with the Tsinghua-Tencent state-of-the-art detectors. \cite{meng} only cites an overall accuracy and recall.}
\label{tencent_compare}
\end{table}

\subsection{Ablation Studies}
In this section we validate the usage of the hierarchical architecture and background threshold. As shown in Table \ref{performance_table}, training a network on both the Bosch and Tsinghua-Tencent dataset resulted in an average performance loss of $27\%$ compared to training a network on only one of the datasets. This result confirms our expectation in Section \ref{introduction} that combining datasets affects performance because it requires the network to approximate a more complex function. We show that this loss is reduced to only $18\%$ when using the hierarchical architecture and background threshold. \\

\noindent\textbf{Effectiveness of Hierarchical Architecture}: We hypothesize the increase in performance from the hierarchical architecture is due to the new loss function being more effective. It is easier to detect the global class of traffic light or traffic sign than to detect the subclasses because there are more training samples of the global class. With a substantial number of global class examples, the network can be trained more efficiently by first detecting the global class then categorizing the detection into local classes.\\

\noindent\textbf{Effectiveness of Background Threshold}: The background threshold also increases the network performance compared to naively training on the combined dataset. This result is expected because the threshold reduces the probability of the RPN selecting non-labelled target objects as the background. There was a concern that requiring the background threshold on a negative image would result in a poor representation of the background, but the increase in performance diminishes this concern. \\

\noindent\textbf{Effectiveness of Hierarchical Architecture and Background Threshold}: The best configuration was the network with both the hierarchical architecture and background threshold. The background threshold and hierarchical architecture benefits are additive as they affect separate parts of the network. We also use this model on a video \cite{la_video} found online and show our model is able to generalize well to traffic lights and signs from a different data source \href{https://www.youtube.com/watch?v=_YmogPzBXOw&feature=youtu.be}{here}. Additional qualitative results are shown in \fig{sample_detection_la}.

\section{Conclusions} \label{conclusion}
It is not common for object detection systems to detect both traffic lights and signs as there are no datasets that contain both traffic light and sign labels, and combining overlapping datasets leads to unlabeled objects of interest. In this paper we bridge this gap by proposing a novel deep hierarchical architecture with a mini-batch selection mechanism that allows our network to train on a combined traffic light and sign dataset. While our network does not achieve the accuracy and recall of certain networks in the literature, it is more suitable for autonomous car deployment. We achieve real-time performance and can perform traffic light and traffic sign detection with one network, thus lowering GPU memory requirements. Further performance could be achieved with more training examples, which could be done through data augmentation.

{\small
\bibliographystyle{unsrt}
\bibliography{traffic_sign_bib}
}

\end{document}